\documentclass{article}
\usepackage{spconf,amsmath,graphicx}
\usepackage{amsfonts}

\usepackage{amssymb}
\usepackage{url}
\usepackage{cite}
\usepackage{hyperref}

\title{Learning Latent Representations for style control and transfer in end-to-end speech synthesis}
\name{Ya-Jie Zhang$^{1 \star }$ \thanks{Paper accepted by IEEE ICASSP 2019} \thanks{$^{\star}$ Work done during internship at Microsoft STC Asia}, Shifeng Pan$^2$, Lei He$^2$, Zhen-Hua Ling$^1$ }

\address{$^1$National Engineering Laboratory for Speech and Language Information Processing,\\ University of Science and Technology of China, Hefei, P.R.China\\ $^2$ Microsoft China\\{\small \tt zyj008@mail.ustc.edu.cn, \{peterpan, helei\}@microsoft.com, zhling@ustc.edu.cn}}

\begin{document}
%
\maketitle
\begin{abstract}
In this paper, we introduce the Variational Autoencoder (VAE) to an end-to-end speech synthesis model, to learn the latent representation of speaking styles in an unsupervised manner.
The style representation learned through VAE shows good properties such as disentangling, scaling, and combination, which makes it easy for style control.
Style transfer can be achieved in this framework by first inferring style representation through the recognition network of VAE, then feeding it into TTS network to guide the style in synthesizing speech.
To avoid Kullback-Leibler (KL) divergence collapse in training, several techniques are adopted.
Finally, the proposed model shows good performance of style control and outperforms Global Style Token (GST) model in ABX preference tests on style transfer.
\end{abstract}
\section{Introduction}
\label{sec:intro}

End-to-end text-to-speech (TTS) models which generate speech directly from characters have made rapid progress in recent years, and achieved very high voice quality \cite{shen2018natural, ping2018deep, li2018close}.
While the single style TTS, usually neutral speaking style, is approaching the extreme quality close to human expert recording \cite{shen2018natural, li2018close}, the interests in expressive speech synthesis also keep rising.
Recently, there also published many promising works in this topic, such as transferring prosody and speaking style within or cross speakers based on end-to-end TTS model \cite{wang2018style, skerry2018towards,stanton2018predicting}.

%
%

Deep generative models, such as Variational Autoencoder (VAE) \cite{kingma2013auto} and Generative Adversarial Network (GAN) \cite{goodfellow2014generative}, are powerful architectures which can learn complicated distribution in an unsupervised manner.
Particularly, VAE, which explicitly models latent variables, have become one of the most popular approaches and achieved significant success on text generation \cite{bowman2015generating}, image generation \cite{higgins2016beta,burgess2018understanding} and speech generation \cite{akuzawa2018expressive,hsu2017learning} tasks.
VAE has many merits, such as learning disentangled factors, smoothly interpolating or continuously sampling between latent representations which can obtain interpretable homotopies \cite{bowman2015generating}.

Intuitively, in speech generation, the latent state of speaker, such as affect and intent, contributes to the prosody, emotion, or speaking style.
For simplicity, we'll hereafter use speaking style to represent these prosody related expressions.
The latent state plays a pretty similar role as the latent variable does in VAE.
Therefore, in this paper we intend to introduce VAE to Tacotron2 \cite{shen2018natural}, a state-of-the-art end-to-end speech synthesis model, to learn the latent representation of speaker state in a continuous space, and further to control the speaking style in speech synthesis.
To be specific, direct manipulation can be easily imposed on the disentangled latent variable, so as to control the speaking style.
On the other hand, with variational inference the latent representation of speaking style can be inferred from a reference audio, which then controls the style of synthesized speech.
Style transfer, from reference audio to synthesized speech, is thus achieved. Last but not least, directly sampling on prior of latent distribution can generate a lot of speech with various speaking style, which is very useful for data augmentation.
Comprehensive evaluation shows the good performance of this method.

We have become aware of recent work by Akuzawa et al. \cite{akuzawa2018expressive} which combines an autoregressive speech
synthesis model with VAE for expressive speech synthesis. The proposed work differs from Akuzawa's as follows:
1) their goal is to synthesize expressive speech, which is achieved by direct sampling from prior of latent distribution at inference stage, while our goal is to control the speaking style of synthesized speech through direct manipulate latent variable or variational inference from a reference audio;
2) the proposed work is on end-to-end TTS model while Akuzawa's not.

The rest of the paper is organized as follows: Section \ref{sec:format} introduces VAE model, our proposed model architecture and tricks for solving KL-divergence collapse problem.
Section \ref{sec:exper} presents the experimental results.
Finally, the paper will be concluded in Section \ref{sec:conc}.

\section{Model}
\label{sec:format}
In this section, we first review Variational Autoencoder.
We then show the details of our proposed style transfer model.

\subsection{Variational Autoencoder}
\label{ssec:vae}

Variational Autoencoder was first defined by Kingma et al. \cite{kingma2013auto} which constructs a relationship between unobserved continuous random latent variables \textbf{z} and observed dataset \textbf{x}.
The true posterior density $p_{\theta}(\textbf{z}|\textbf{x})$ is intractable, which results in an indifferentiable marginal likelihood $p_{\theta}(\textbf{x})$.
To address this, a recognition model $q_{\phi}(\textbf{z}|\textbf{x})$ is introduced as an approximation to the intractable posterior.
Following the variational principle, $\log p_{\theta}(\textbf{x})$ can be rewritten as shown in equation (\ref{f1}), where $\mathcal{L}(\theta,\phi;\textbf{x})$ is the variational lower bound to optimize.
\begin{equation}
\begin{aligned}
\log p_{\theta}(\textbf{x}) &= KL[q_{\phi}(\textbf{z}|\textbf{x})||p_{\theta}(\textbf{z}|\textbf{x})] + \mathcal{L}(\theta,\phi;\textbf{x})
\\& \geq \mathcal{L}(\theta,\phi;\textbf{x})
\\&=\mathbb{E}_{q_{\phi}(\textbf{z}|\textbf{x})}[\log p_{\theta}(\textbf{x}|\textbf{z})] - KL[q_{\phi}(\textbf{z}|\textbf{x})||p_{\theta}(\textbf{z})]
\end{aligned}
\label{f1}	
\end{equation}

Generally, the prior over latent variables $p_{\theta}(\textbf{z})$ is assumed to be centered isotropic multivariate Gaussian $\mathcal{N}(\textbf{z};\textbf{0},\textbf{I})$, where \textbf{I} is the identity matrix. The usual choice of $q_{\phi}(\textbf{z}|\textbf{x})$ is $\mathcal{N}(\textbf{z};\boldsymbol{\mu}(\textbf{x}),\boldsymbol{\sigma}^2(\textbf{x})\textbf{I})$, so that $KL[q_{\phi}(\textbf{z}|\textbf{x})||p_{\theta}(\textbf{z})]$ can be calculated in closed form.
In practice, $\boldsymbol{\mu}(\textbf{x})$ and $\boldsymbol{\sigma}^2(\textbf{x})$ are learned from observed dataset via neural networks which can be viewed as an encoder.
The expectation term in equation (\ref{f1}) plays the role of decoder which decodes latent variables \textbf{z} to reconstruct \textbf{x}.
The decoder may produce the expected reconstruction if the output of decoder is averaged over many samples of \textbf{x} and \textbf{z} \cite{doersch2016tutorial}.
In the rest of the paper, $-\mathbb{E}_{q_{\phi}(\textbf{z}|\textbf{x})}[\log p_{\theta}(\textbf{x}|\textbf{z})]$ is referred to as reconstruction loss and $KL[q_{\phi}(\textbf{z}|\textbf{x})||p_{\theta}(\textbf{z})]$ is referred to as KL loss.


Stochastic inputs can be processed by stochastic gradient descent via backpropagation, but stochastic units within the network cannot be processed by backpropagation.
Thus, in practice, "reparameterization trick" is introduced to VAE framework.
Sampling \textbf{z} from distribution $\mathcal{N}(\boldsymbol{\mu},\boldsymbol{\sigma}^2\textbf{I})$ is decomposed to first sampling $\boldsymbol{\epsilon} \sim \mathcal{N}(\textbf{0},\textbf{I})$ and then computing $\textbf{z}=\boldsymbol{\mu}+\boldsymbol{\sigma}\odot\boldsymbol{\epsilon}$, where $\odot$ denotes an element-wise product.

\subsection{Proposed Model Architecture}
\label{ssec:archi}

\begin{figure}[t]
\centering
\includegraphics[width=7.5cm]{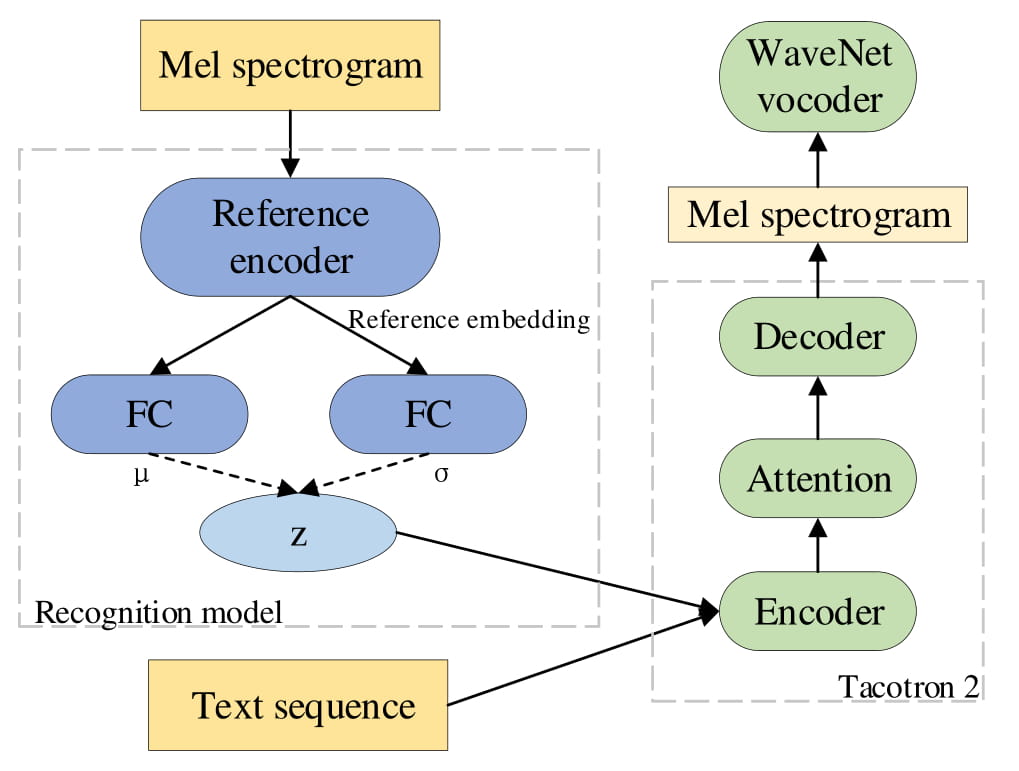}
\caption{\label{fig:FIG1}  An architecture of the proposed style transfer TTS model. The dashed lines denote sampling \textbf{z} from parametric distribution.}

\end{figure}

In this work, we introduce VAE into end-to-end TTS model and propose a flexible model for style control and style transfer.
The whole network consists of two components, as shown in Fig.\ref{fig:FIG1}: (1) A recognition model or inference network which encodes reference audio into a fixed-length short vector of latent representation (or latent variables \textbf{z} which stand for style representation), and (2) an end-to-end TTS model based on Tacotron 2, which converts the combined encoder states (including latent representations and text encoder states) to generated target sentence with specific style.

The input texts are character sequences and the acoustic features are mel-frequency spectrograms.
One may use various powerful and complex neural networks for the recognition model.
Here, we only adopt a recurrent reference encoder followed by two fully connected layers.
We use the same architecture and hyperparameters for reference encoder as Wang et al. \cite{wang2018style} which consists of six 2-D convolutional layers followed by a GRU layer.
The output, which denotes some embedding of the reference audio, is then passed through two separate fully connected (FC) layers with linear activation function to generate the mean and standard deviation of latent variables \textbf{z}.
The prior and approximative posterior are Gaussian distribution mentioned Section \ref{ssec:vae}.
Then \textbf{z} is derived by reparameterization trick.
The encoder which deals with character inputs consists of three 1-D convolutional layers with 5 width and 512 channels followed by a bidirectional \cite{schuster1997bidirectional} LSTM \cite{hochreiter1997long} layer using zoneout \cite{krueger2016zoneout} with probability 0.1.
The output text encoder state is simply added by \textbf{z} and then is consumed by a location-sensitive attention network \cite{chorowski2015attention} which converts encoded sequence to a fixed-length context vector for each decoder output step.
In addition, \textbf{z} should be first passed through a FC layer to make sure the dimension equal to text encoder state before add operation.
The attention module and decoder have the same architecture as Tacotron 2 \cite{shen2018natural}.
Then, WaveNet \cite{van2016wavenet} vocoder is utilized to reconstruct waveform.

The total loss of proposed model is shown in equation (\ref{f2}).
\begin{equation}
\begin{aligned}
Loss =KL[q_{\phi}(\textbf{z}|\textbf{x})||p_{\theta}(\textbf{z})]-\mathbb{E}_{q_{\phi}(\textbf{z}|\textbf{x})}[\log p_{\theta}(\textbf{x}|\textbf{z},\textbf{t})]+l_{stop}
\end{aligned}
\label{f2}	
\end{equation}
Compared with the lower bound in equation (\ref{f1}), the reconstruction loss term is conditioned on both latent variable \textbf{z} and input text \textbf{t} and a stop token loss $l_{stop}$ is added.
It is worth mentioning that, after comparing L2-loss with negative log likelihood of Gaussian distribution, we finally choose L2-loss of mel spectrograms as reconstruction loss. 

\subsection{Resolve KL collapse problem}
\label{ssec:kl}
During training, we observe that the KL loss $KL[q_{\phi}(\textbf{z}|\textbf{x})||p_{\theta}(\textbf{z})]$ is always found collapsed before they learned a distinguishable representation, which is a common phenomenon but a crucial issue in training VAE models.
In other words, the convergence speed of KL loss far surpasses that of the reconstruction loss and the KL loss quickly drops to nearly zero and never rises again, which means the encoder doesn't work.
Thus, KL annealing \cite{bowman2015generating} is introduced to our task to solve this problem.
That is, during training, add a variable weight to the KL term.
The weight is close to zero at the beginning of training and then  gradually increase.
In addition, KL loss is taken into account once every K steps.
By combining these two tricks, the KL loss keeps nonzero and avoids to collapse.

\section{Experiments and Analysis}
\label{sec:exper}

\begin{figure}[t]
\centering
\includegraphics[width=8cm]{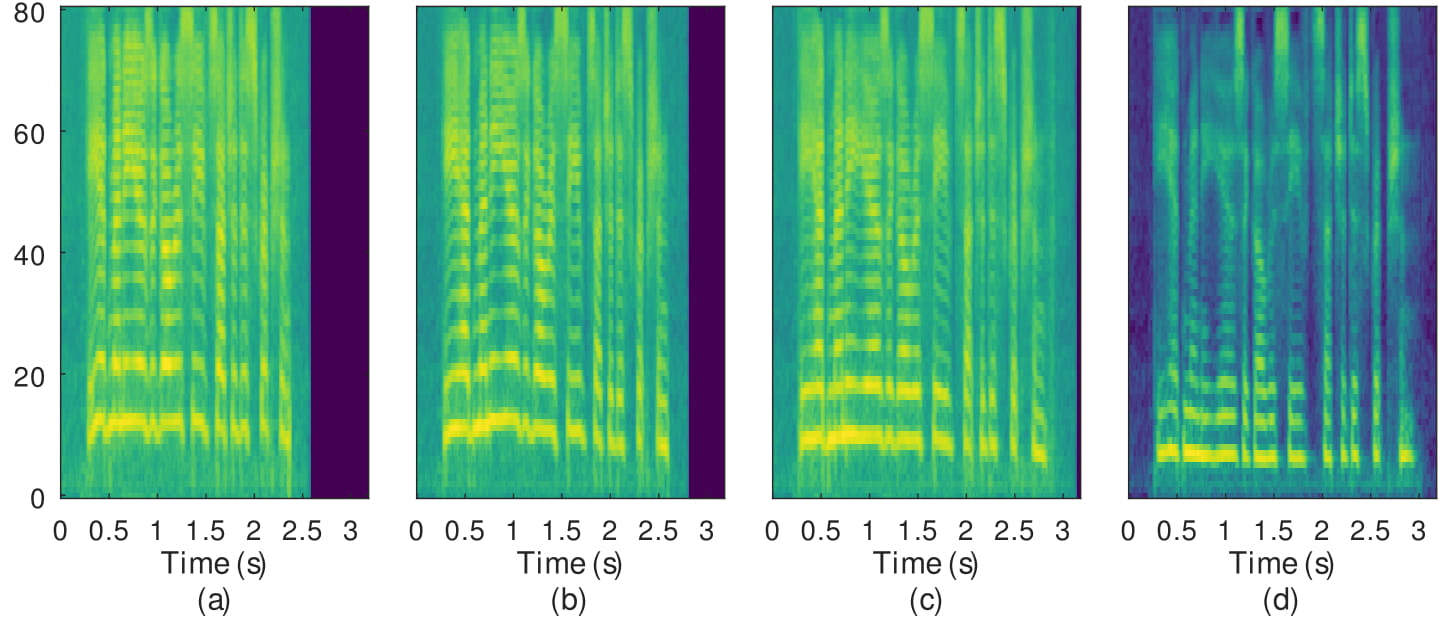}
\caption{\label{fig:FIG2} Spectrograms generated by interpolation between two \textbf{z}. The interpolation coefficient is : (a) $\textbf{z}_a$, (b) $\frac{1}{3}\textbf{z}_a+\frac{2}{3}\textbf{z}_d$, (c) $\frac{2}{3}\textbf{z}_a+\frac{1}{3}\textbf{z}_d$, (d) $\textbf{z}_d$ .}
\end{figure}

\begin{figure}[t]
\centering
\includegraphics[width=8cm]{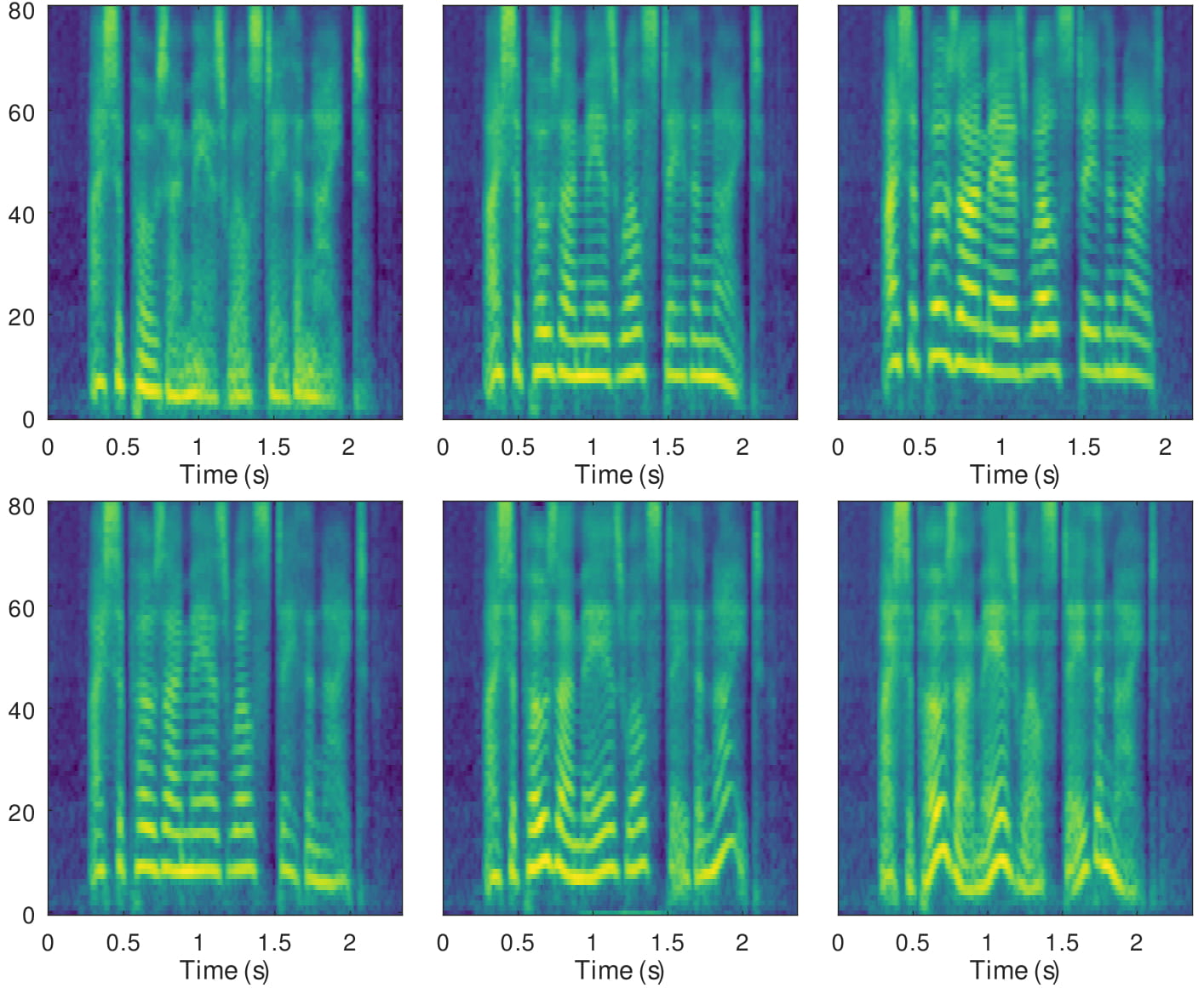}
\caption{\label{fig:FIG3}Spectrograms generated to demonstrate disentangled factors. The first row exhibits the control of pitch height only by adjusting latent dimension 6 to be -0.9, -0.1, 0.7. The second row shows that the local pitch variation is gradually magnified by increasing the value of dimension 10, which is 0.1, 0.5, 0.9, respectively.
}
\end{figure}

\subsection{experimental setup}
\label{sec:confi}
An 105-hour audiobook recordings dataset read with various storytelling styles by a single English speaker (Blizzard Challenge 2013) was used in our experiments.
The dataset contains 58453 utterances for training and 200 for test.
80-dimensional mel spectrograms were extracted with frame shift 12.5 ms and frame length 50 ms.
GST model \cite{wang2018style} with character inputs was used as our baseline model.
The hyperparameters are set according to \cite{wang2018style}.
As for our proposed model, the dimension of latent variables is 32.
The parameter K mentioned in \ref{ssec:kl} is 100 before 15000 training steps and 400 after the threshold.


At inference stage, in evaluation of style control, we directly manipulate \textbf{z} without going through the whole recognition model.
With regard to evaluation of style transfer, we feed audio clips as reference and go through the recognition model.
Both parallel and non-parallel style transfer audios are generated and evaluated\footnote{The audio samples can be found at \url{http://home.ustc.edu.cn/~zyj008/ICASSP2019}.}.
Parallel transfer means the target text information is the same as reference audio's, vice versa.

\subsection{Style control}

\subsubsection{Interpolation of latent variables}

As mentioned in \cite{bowman2015generating}, VAE supports smoothly interpolation and continuous sampling between latent representations, which obtains interpretable homotopies.
Thus, we did interpolation operation between two \textbf{z}\footnote{These two \textbf{z} are derived by feeding two audios to the recognition model, one with high speaking rate and high-pitch, the other with low speaking rate and low-pitch.}.
One can generate speech with high speaking rate and high-pitch, the other with low speaking rate and low-pitch.
The mel spectrograms of generated speech are shown in Fig. \ref{fig:FIG2}.
As we can see, both pitch and speaking rate of generated speech gradually decrease along with the interpolating.
The result shows that the learnt latent space is continuous in controlling the trend of spectrograms which will further reflect in the change of style.

\subsubsection{Disentangled factors}

A disentangled representation means that a latent variable completely controls a concept alone and is invariant to changes from other factors \cite{higgins2016beta}.
In experiments, we found that several dimensions of \textbf{z} could independently control different style attributes, such as pitch-height, local pitch variation, speaking rate.
Fig. \ref{fig:FIG3} shows the alteration of spectrograms by manipulating single dimension while fixing others.
Adjusting one of these dimensions, only one attribute of generated speech changes.
This shows that, in our model, VAE has the ability of learning disentangled latent factors.

\begin{figure}[t]
\centering
\includegraphics[width=8cm]{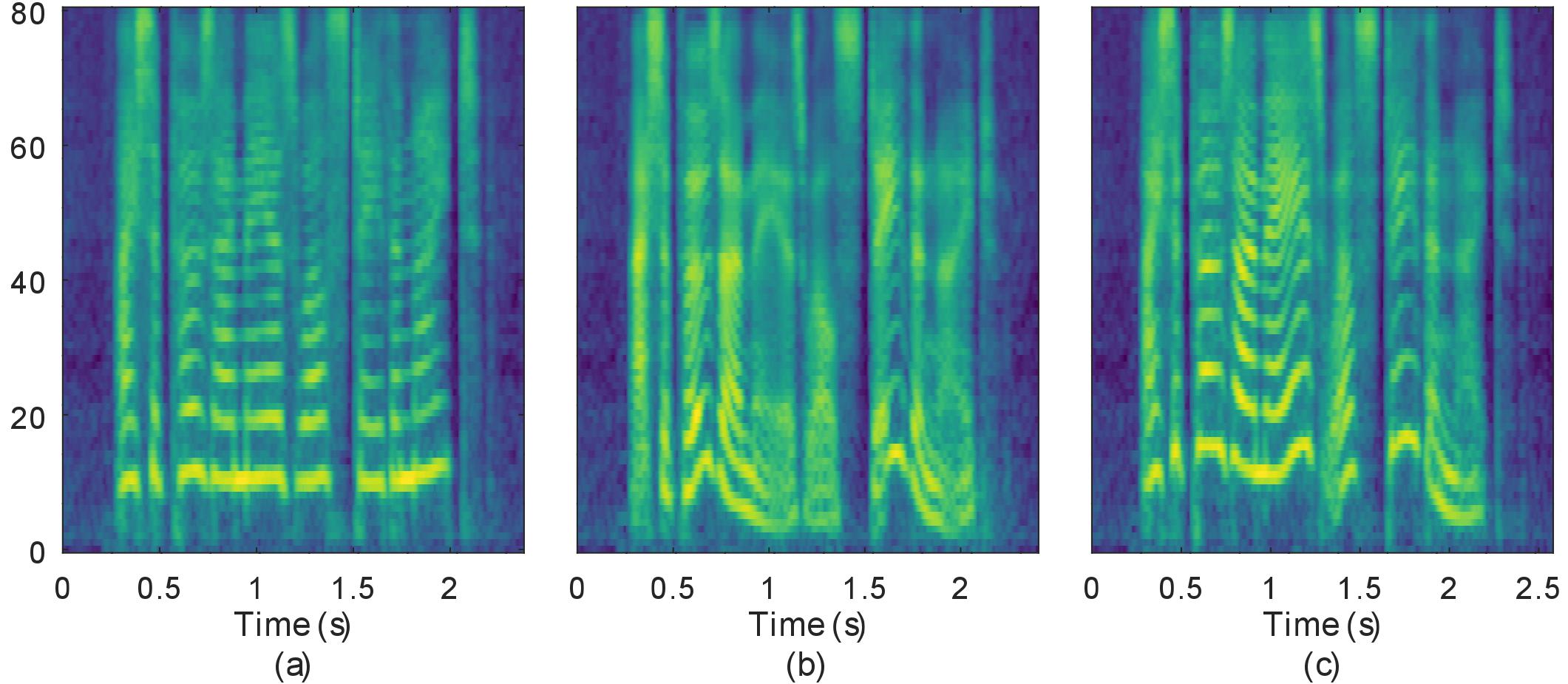}
\caption{\label{fig:FIG4}Audio (a) and (b) are generated with \textbf{z} which setting a single dimension to be non-zero with other dimensions to be zero. The valued dimension in (a) controls pitch height, while in (b) controls pitch variation. (c) is generated with the summation of \textbf{z} from (a) and (b). }
\end{figure}

Next, we combined two disentangled dimensions to verify the additivity of latent variables.
Fig. \ref{fig:FIG4} illustrates the combination results of pitch height and local pitch variation attributes.
It shows that the audio generated with combined \textbf{z} inherits the characteristics of both disentangled dimensions.

\subsection{Style transfer}

\begin{figure}[t]
\centering
\includegraphics[width=8cm]{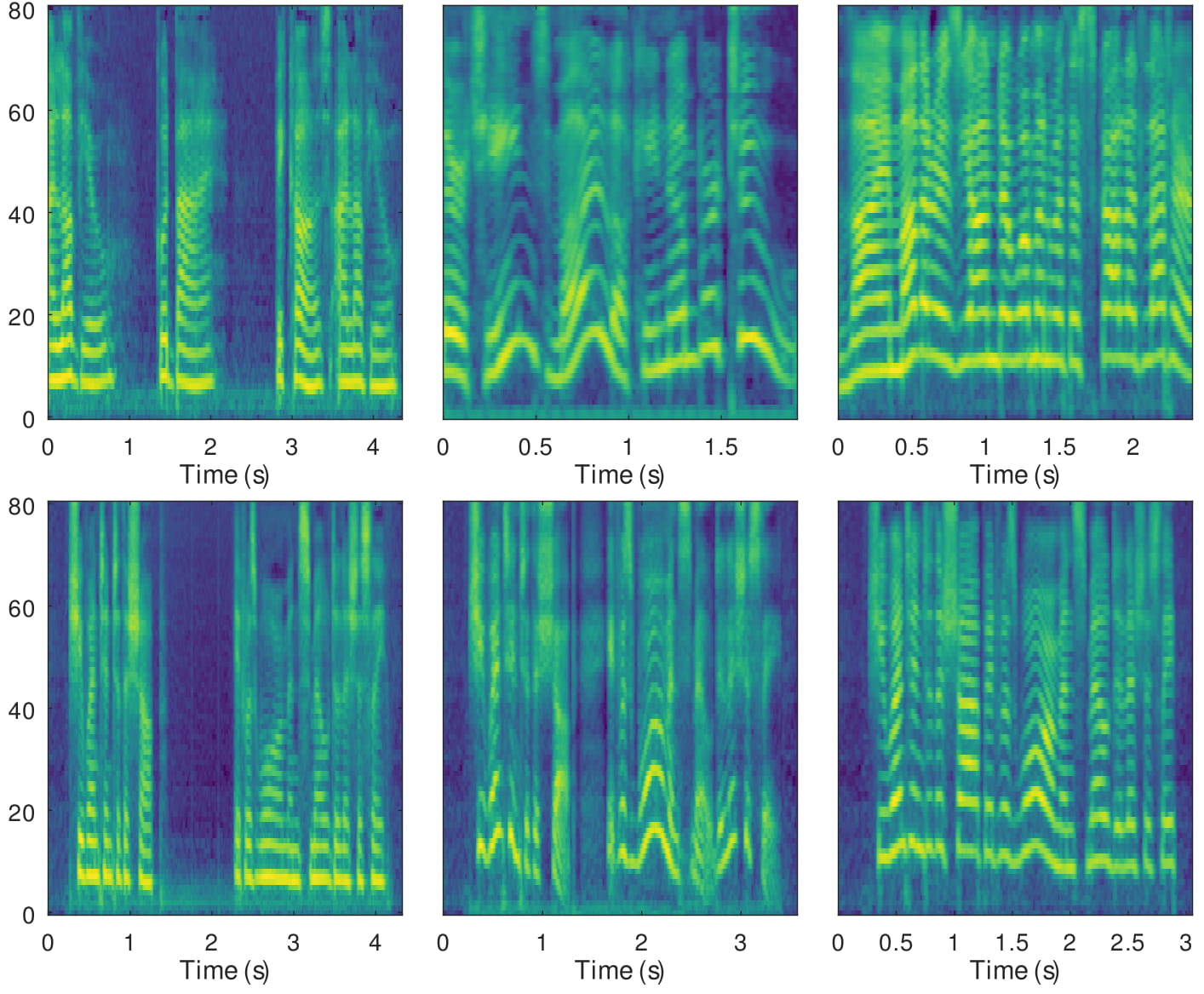}
\caption{\label{fig:FIG5}The first row exhibits the mel spectrograms of three recordings with different styles, while the second row exhibits the synthesised audios referenced on those recordings separately. The synthesised audios have the same text "She went into the shop . It was warm and smelled deliciously."}
\end{figure}

Fig. \ref{fig:FIG5} shows mel spectrograms of the style transferred synthetic speech aligned with their corresponding references.
The reference audios are chosen from test set with certain styles.
The synthesized audios share the same input text.
As we can see in Fig. \ref{fig:FIG5}, the mel spectrograms of generated speech and their reference audio have pattern similarities, such as in pitch-height, pause time, speaking rate and pitch variation.

\begin{figure}[t]
\centering
\includegraphics[width=8cm]{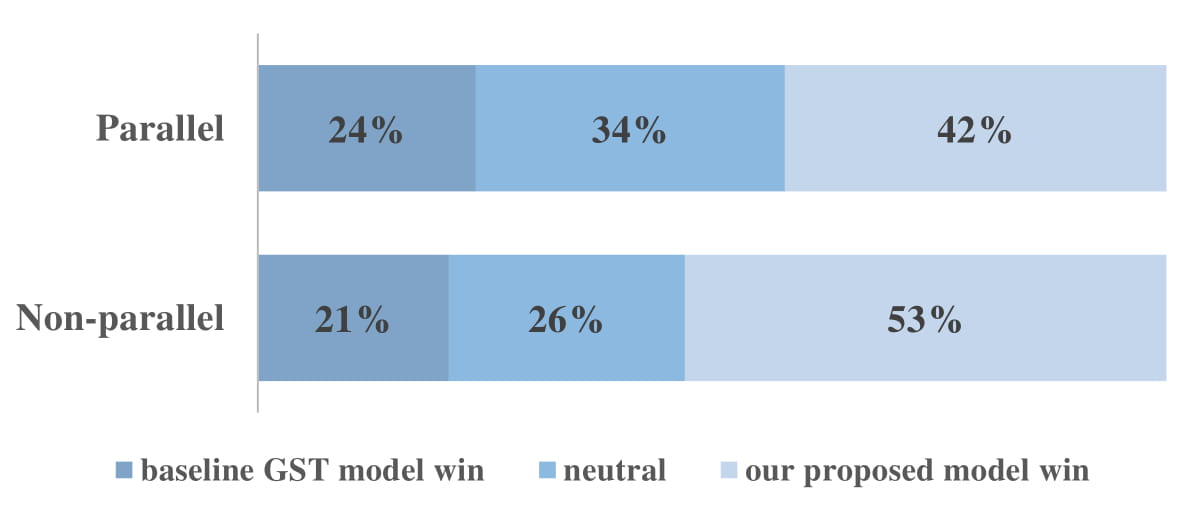}
\caption{\label{fig:FIG6}ABX test results for parallel and non-parallel transfers.}
\end{figure}

\subsection{Subjective test}

To subjectively evaluate the performance of style transfer, crowd-sourcing ABX preference tests on parallel and non-parallel transfer were conducted.
For parallel transfer, 60 audio clips with their texts are randomly selected from test set.
For non-parallel transfer, 60 sentences of text and 60 other reference audio clips are selected to generate speech.
The baseline voice is generated from the best GST model we have built.
Each case in ABX test is judged by 25 native judgers.
The total number of judger is 56 for parallel test and 57 for non-parallel test.
The criterion in rating is "which one's speaking style is closer to the reference style" with three choices: (1) 1st is better; (2) 2nd is better; (3) neutral.

Fig. \ref{fig:FIG6} shows the ABX results.
As we can see, the proposed model outperforms GST model on both parallel and non-parallel style transfer (at p-value $<$ 10$^{-5}$).
It shows that VAE can better model the latent style representations, which results in better style transfer.
What's more, the performance of the proposed model on non-parallel transfer is much better than that on parallel transfer, which shows the better generalization capability of the proposed model.

\section{Conclusion}
\label{sec:conc}
A VAE module is introduced to end-to-end TTS model, to learn the latent representation of speaking style in a continuous space in an unsupervised manner, which then can control the speaking style in synthesized speech.
We have demonstrated that the latent space is continuous and explored the disentangled factors in learned latent variables.
The proposed model shows good performance in style transfer, which outperforms GST model via ABX test, especially in non-parallel transfer.

Future work will keep focusing on getting better disentangled and interpretable latent representations. In addition, the scope of style transfer research will further extend to multi-speakers, instead of single speaker.

\bibliographystyle{IEEEbib}
\bibliography{strings,refs}

\begin{thebibliography}{10}

\bibitem{shen2018natural}
Jonathan Shen, Ruoming Pang, Ron~J Weiss, Mike Schuster, Navdeep Jaitly,
  Zongheng Yang, Zhifeng Chen, Yu~Zhang, Yuxuan Wang, Rj~Skerrv-Ryan, et~al.,
\newblock ``Natural {TTS} synthesis by conditioning {WaveNet} on mel
  spectrogram predictions,''
\newblock in {\em 2018 IEEE International Conference on Acoustics, Speech and
  Signal Processing (ICASSP)}. IEEE, 2018, pp. 4779--4783.

\bibitem{ping2018deep}
Wei Ping, Kainan Peng, Andrew Gibiansky, Sercan~O Arik, Ajay Kannan, Sharan
  Narang, Jonathan Raiman, and John Miller,
\newblock ``Deep voice 3: Scaling text-to-speech with convolutional sequence
  learning,''
\newblock 2018.

\bibitem{li2018close}
Naihan Li, Shujie Liu, Yanqing Liu, Sheng Zhao, Ming Liu, and Ming Zhou,
\newblock ``Close to human quality {TTS} with transformer,''
\newblock {\em arXiv preprint arXiv:1809.08895}, 2018.

\bibitem{wang2018style}
Yuxuan Wang, Daisy Stanton, Yu~Zhang, RJ~Skerry-Ryan, Eric Battenberg, Joel
  Shor, Ying Xiao, Fei Ren, Ye~Jia, and Rif~A Saurous,
\newblock ``Style tokens: Unsupervised style modeling, control and transfer in
  end-to-end speech synthesis,''
\newblock in {\em Proceedings of the 35th International Conference on Machine
  Learning, {ICML} 2018,}, 2018, pp. 5167--5176.

\bibitem{skerry2018towards}
RJ~Skerry-Ryan, Eric Battenberg, Ying Xiao, Yuxuan Wang, Daisy Stanton, Joel
  Shor, Ron~J Weiss, Rob Clark, and Rif~A Saurous,
\newblock ``Towards end-to-end prosody transfer for expressive speech synthesis
  with {Tacotron},''
\newblock in {\em Proceedings of the 35th International Conference on Machine
  Learning,{ICML} 2018}, 2018, pp. 4700--4709.

\bibitem{stanton2018predicting}
Daisy Stanton, Yuxuan Wang, and RJ~Skerry-Ryan,
\newblock ``Predicting expressive speaking style from text in end-to-end speech
  synthesis,''
\newblock {\em arXiv preprint arXiv:1808.01410}, 2018.

\bibitem{kingma2013auto}
Diederik~P Kingma and Max Welling,
\newblock ``Auto-encoding variational bayes,''
\newblock in {\em Proc. 2nd International Conference on Learning
  Representations}, 2014.

\bibitem{goodfellow2014generative}
Ian Goodfellow, Jean Pouget-Abadie, Mehdi Mirza, Bing Xu, David Warde-Farley,
  Sherjil Ozair, Aaron Courville, and Yoshua Bengio,
\newblock ``Generative adversarial nets,''
\newblock in {\em Advances in neural information processing systems}, 2014, pp.
  2672--2680.

\bibitem{bowman2015generating}
Samuel~R Bowman, Luke Vilnis, Oriol Vinyals, Andrew~M Dai, Rafal Jozefowicz,
  and Samy Bengio,
\newblock ``Generating sentences from a continuous space,''
\newblock in {\em Proceedings of the 20th {SIGNLL} Conference on Computational
  Natural Language Learning, CoNLL 2016}.

\bibitem{higgins2016beta}
Irina Higgins, Loic Matthey, Arka Pal, Christopher Burgess, Xavier Glorot,
  Matthew Botvinick, Shakir Mohamed, and Alexander Lerchner,
\newblock ``$\beta$-{VAE}: Learning basic visual concepts with a constrained
  variational framework,''
\newblock 2016.

\bibitem{burgess2018understanding}
Christopher~P Burgess, Irina Higgins, Arka Pal, Loic Matthey, Nick Watters,
  Guillaume Desjardins, and Alexander Lerchner,
\newblock ``Understanding disentangling in $\beta$-{VAE},''
\newblock {\em arXiv preprint arXiv:1804.03599}, 2018.

\bibitem{akuzawa2018expressive}
Kei Akuzawa, Yusuke Iwasawa, and Yutaka Matsuo,
\newblock ``Expressive speech synthesis via modeling expressions with
  variational autoencoder,''
\newblock in {\em Proc. Interspeech 2018}, 2018, pp. 3067--3071.

\bibitem{hsu2017learning}
Wei-Ning Hsu, Yu~Zhang, and James Glass,
\newblock ``Learning latent representations for speech generation and
  transformation,''
\newblock in {\em Proc. Interspeech 2017}, 2017, pp. 1273--1277.

\bibitem{doersch2016tutorial}
Carl Doersch,
\newblock ``Tutorial on variational autoencoders,''
\newblock {\em arXiv preprint arXiv:1606.05908}, 2016.

\bibitem{schuster1997bidirectional}
Mike Schuster and Kuldip~K Paliwal,
\newblock ``Bidirectional recurrent neural networks,''
\newblock {\em IEEE Transactions on Signal Processing}, vol. 45, no. 11, pp.
  2673--2681, 1997.

\bibitem{hochreiter1997long}
Sepp Hochreiter and J{\"u}rgen Schmidhuber,
\newblock ``Long short-term memory,''
\newblock {\em Neural computation}, vol. 9, no. 8, pp. 1735--1780, 1997.

\bibitem{krueger2016zoneout}
David Krueger, Tegan Maharaj, J{\'a}nos Kram{\'a}r, Mohammad Pezeshki, Nicolas
  Ballas, Nan~Rosemary Ke, Anirudh Goyal, Yoshua Bengio, Aaron Courville, and
  Chris Pal,
\newblock ``Zoneout: Regularizing rnns by randomly preserving hidden
  activations,''
\newblock in {\em Proc. ICLR,2017}, 2017.

\bibitem{chorowski2015attention}
Jan~K Chorowski, Dzmitry Bahdanau, Dmitriy Serdyuk, Kyunghyun Cho, and Yoshua
  Bengio,
\newblock ``Attention-based models for speech recognition,''
\newblock in {\em Advances in neural information processing systems}, 2015, pp.
  577--585.

\bibitem{van2016wavenet}
A{\"a}ron Van Den~Oord, Sander Dieleman, Heiga Zen, Karen Simonyan, Oriol
  Vinyals, Alex Graves, Nal Kalchbrenner, Andrew~W Senior, and Koray
  Kavukcuoglu,
\newblock ``{WaveNet}: A generative model for raw audio.,''
\newblock in {\em SSW}, 2016, p. 125.

\end{thebibliography}

\end{document}